%% file: main.tex
\documentclass[10pt,twocolumn,letterpaper, table]{article}

\usepackage{iccv}
\usepackage{times}
\usepackage{epsfig}
\usepackage{graphicx}
\usepackage{amsmath}
\usepackage{amssymb}

\iccvfinalcopy 

\ificcvfinal
\def\assignedStartPage{9876} 
\fi

\ificcvfinal
\usepackage[breaklinks=true,bookmarks=false]{hyperref}
\else
\usepackage[pagebackref=true,breaklinks=true,colorlinks,bookmarks=false]{hyperref}
\fi

\ificcvfinal
\else
\fi

\input{macros}

\begin{document}
\title{Video Manipulations Beyond Faces: A Dataset with Human-Machine Analysis}

\author{Trisha Mittal \thanks{Work done as an intern at Adobe Research.}\\
University of Maryland\\
{\tt\small trisha@umd.edu}
\and
Ritwik Sinha\\
Adobe Research, San Jose\\
{\tt\small risinha@adobe.com}
\and
Viswanathan Swaminathan\\
Adobe Research, San Jose\\
{\tt\small vishy@adobe.com}
\and
John Collomosse\\
Adobe Research, San Jose\\
{\tt\small collomos@adobe.com}
\and
Dinesh Manocha\\
University of Maryland\\
{\tt\small dmanocha@umd.edu}
}

\maketitle

\input{sections/0-abstract}
\input{sections/1-introduction}
\input{sections/2-relatedwork}
\input{sections/3-dataset}
\input{sections/4-experiments}
\input{sections/7-discussion}
{\footnotesize
\bibliographystyle{plain}
\bibliography{refs}}
\clearpage
\appendix
\input{sections/appendix.tex}
\end{document}

%% file: macros.tex
\usepackage{stfloats}
\usepackage{xcolor}
\usepackage{blindtext}
\usepackage{hyperref}
\usepackage{graphicx}
\usepackage{amsmath}
\usepackage{booktabs}
\usepackage{threeparttable}
\usepackage{multirow}
\usepackage{multicol}
\usepackage{subcaption}
\usepackage{url}
\usepackage{tabularx}
\usepackage{amsfonts}
\usepackage{balance}
\usepackage{pifont}
\usepackage[font=small]{caption}
\usepackage{float}
\newcommand{\cmark}{\ding{51}}
\newcommand{\xmark}{\ding{55}}

\newcommand{\dataname}{\textsc{VideoSham}}
\usepackage{bm}
\usepackage{soul}
\usepackage[export]{adjustbox}
\usepackage{array}
\usepackage[neverdecrease]{paralist}
\newcommand\MyBox[2]{
  \fbox{\lower0.75cm
    \vbox to 1.7cm{\vfil
      \hbox to 1.7cm{\hfil\parbox{1.4cm}{#1\\#2}\hfil}
      \vfil}%
  }%
}
\newcommand\Tstrut{\rule{0pt}{2.6ex}}         
\newcommand{\rulesep}{\unskip\ \vrule\ }

\definecolor{sg}{HTML}{00ff7f}
\definecolor{lb}{HTML}{b0f5ef}
\definecolor{lg}{HTML}{9bfaa8}

%% file: sections/0-abstract.tex
\begin{abstract}
As tools for content editing mature, and artificial intelligence (AI) based algorithms for synthesizing media grow, the presence of manipulated content across online media is increasing. This phenomenon causes the spread of misinformation, creating a greater need to distinguish between ``real'' and ``manipulated'' content. To this end, we present \dataname, a dataset consisting of $826$ videos ($413$ real and $413$ manipulated). Many of the existing deepfake datasets focus exclusively on two types of facial manipulations---swapping with a different subject's face or altering the existing face. \dataname, on the other hand, contains more diverse, context-rich, and human-centric, high-resolution videos manipulated using a combination of $6$ different spatial and temporal attacks. Our analysis shows that state-of-the-art manipulation detection algorithms only work for a few specific attacks and do not scale well on \dataname. We performed a user study on Amazon Mechanical Turk with $1200$ participants to understand if they can differentiate between the real and manipulated videos in \dataname. Finally, we dig deeper into the strengths and weaknesses of performances by humans and SOTA-algorithms to identify gaps that need to be filled with better AI algorithms. We present the dataset here\footnote{\href{https://github.com/adobe-research/VideoSham-dataset}{\dataname~ dataset link.}}.
\end{abstract}

%% file: sections/1-introduction.tex
\section{Introduction}
\input{images/example}
\input{tables/summary}

\input{tables/attacks}
The proliferation of accessible video editing software and artificial intelligence~(AI) tools has led to an increase in manipulated video content~\cite{khelifi2017perceptual,he2021forgerynet}. While digital manipulation is commonplace in the creative process, in some cases video manipulation has a malicious intent. Social media often amplifies such false information through the circulation of manipulated videos~\cite{anderson2018,alvaro2017}. A recent survey by Pew Research Center showed that exposure to such false information is of widespread concern~\cite{silver_2020}. Therefore, there has been a significant increase in cases of misinformation, fraud and cybercrimes in the last decade. Such video manipulations pose a great threat to politics and can manipulate elections~\cite{watts2021measuring,allen2020evaluating}, alter political narratives, weaken the public's trust in a country's leadership, and an increase hatred among various social groups. Another common occurrence is corporate frauds and scams where people use altered audio to impersonate other people to extort cash and other resources. Lastly, many video manipulations often result in numerous cybercrimes~\cite{harris2018deepfakes,spivak2018deepfakes,botha2020fake}. To further illustrate our motivations in this work, we depict such instances of video manipulations in Figure~\ref{fig:motivation}.

This leads to an important question---how do we detect manipulated content? The current arsenal of techniques involve the use of AI which in turn requires tremendous amounts of data. In the past decade alone, there has been a surge in the number of benchmark deepfake datasets~\cite{DeepFake-TIMIT, faceforensics++, DFDC} which manipulate the facial features of subjects in images and videos. We summarize recent deepfake datasets in Table~\ref{tab:summary}. 

But facial manipulations represent \textit{only a fraction} of all manipulated content circulated on social media. For example, modifications also include changing the background context~(Figure~\ref{fig:motivation}b), text and audio~(Figure~\ref{fig:motivation}c) in media, aesthetic edits, adding/removing entities~(Figure~\ref{fig:motivation}a), and temporal edits~(Figure~\ref{fig:motivation}d). These manipulations can be performed in a matter of clicks due to the availability of state of the art video editing tools like Adobe AfterEffects$^{\text{TM}}$, Adobe Lightroom$^{\text{TM}}$, Filmora, GIMP, and many others. To our knowledge, no benchmark video dataset exists that extends beyond deepfake-only facial manipulations to include the vast range of manipulations described above.

\subsection*{Main Contributions}
We release a new manipulated high-resolution video dataset called \dataname~(Figure~\ref{fig:qual}). \dataname~offers the following benefits over existing manipulated video datasets:
\begin{enumerate}
    \item Beyond Faces (Deepfakes): The videos in \dataname~are manipulated using six spatial and temporal attacks (See Table~\ref{tab:attacks}) manipulating videos at the scene level targeting, not just faces, but also the background context, text and audio, aesthetic edits, adding/removing entities, and temporal edits (See Figure~\ref{fig:cover}).
    
    \item Beyond Images: Although there exist image manipulation datasets that go beyond faces, they cannot be used to detect video manipulations, which require dedicated video datasets. The latter, however, are hard to create due to the manual labor involved. In this work, we go beyond images to release the first video manipulation dataset containing beyond-face manipulations.
\end{enumerate}

\noindent \dataname~consists of $413$ real-world videos and their corresponding manipulated versions~(total $826$ videos). The videos have diverse scene backgrounds, are context-rich, and contain up to $9$ subjects on average. \dataname~is the largest dataset containing manipulated videos generated by professional video editors with varied attacks. A user study conducted on Amazon Mechanical Turk~(AMT) to understand the kind of attack methods that mislead humans the most. In addition, we analyze the performance of existing state of the art deepfake detection algorithms and video forensics algorithms on \dataname. We find that these techniques are less than $50\%$ effective in distinguishing between a real and a manipulated video. 

We elaborate more about \dataname~in Section~\ref{sec:dataset}. In Section~\ref{sec:experiments} we present our findings from the user study and evaluation of detection models. And, finally in Section~\ref{sec:discussions}, we discuss some promising ideas to help these attacks. 

%% file: images/example.tex
\begin{figure*}[t]
\centering
\begin{adjustbox}{minipage=\linewidth,scale=1}
\begin{subfigure}{0.32\textwidth}
\includegraphics[width=\linewidth]{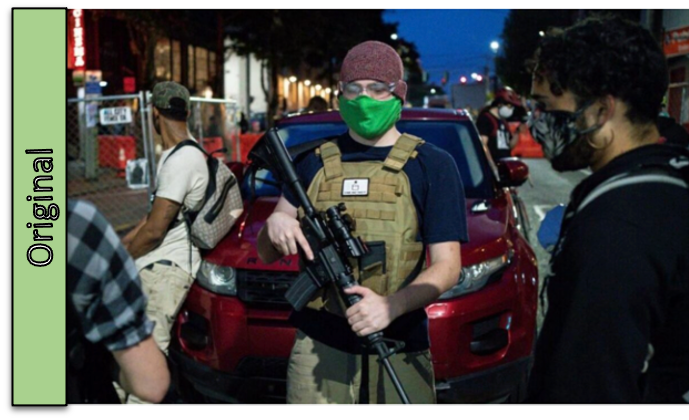}
\caption*{\textbf{(a1)} The original photo, from Getty Images shows an armed man parked in front of a car. }
\end{subfigure}
\hfill
\rulesep
\begin{subfigure}{0.32\textwidth}
\includegraphics[width=\linewidth]{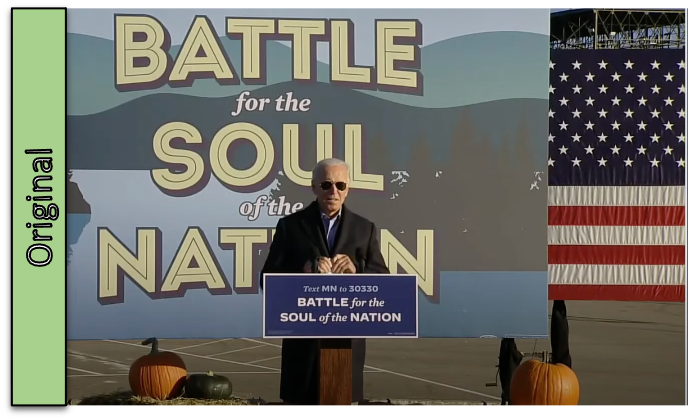}
\caption*{\textbf{(b1)} This is an original clip of a presidential candidate addressing public in the US state, Minnesota.}
\end{subfigure}
\hfill
\rulesep
\begin{subfigure}{0.32\textwidth}
\includegraphics[width=\linewidth]{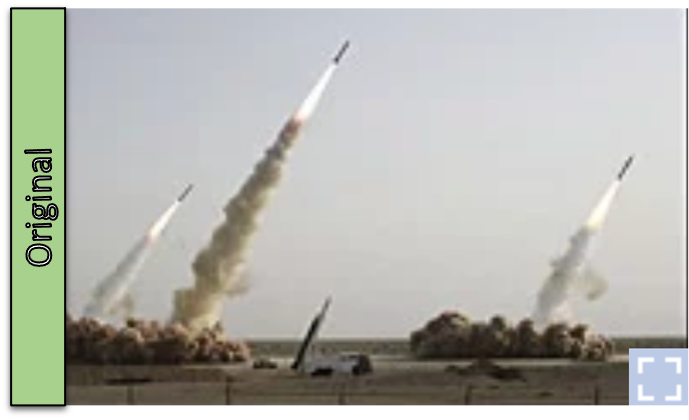}
\caption*{\textbf{(c1)} An original image shows three missiles being launched by Iran's government.}
\end{subfigure}
\begin{subfigure}{0.32\textwidth}
\includegraphics[width=\linewidth]{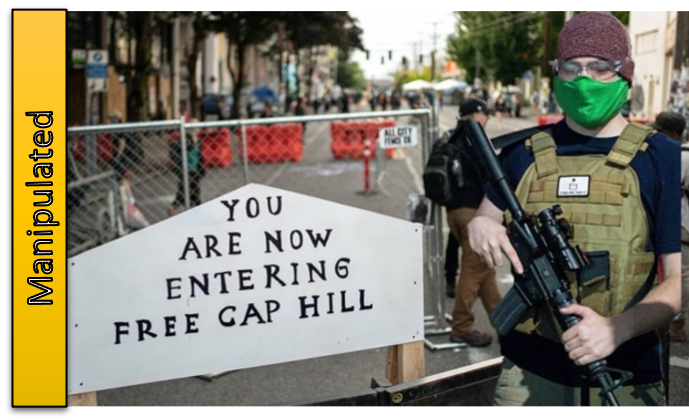}
\caption*{\textbf{(a2)} The photo above was altered by digitally placing the armed man in front of a peaceful protest, insinuating violence. }
\end{subfigure}
\hfill
\rulesep
\begin{subfigure}{0.32\textwidth}
\includegraphics[width=\linewidth]{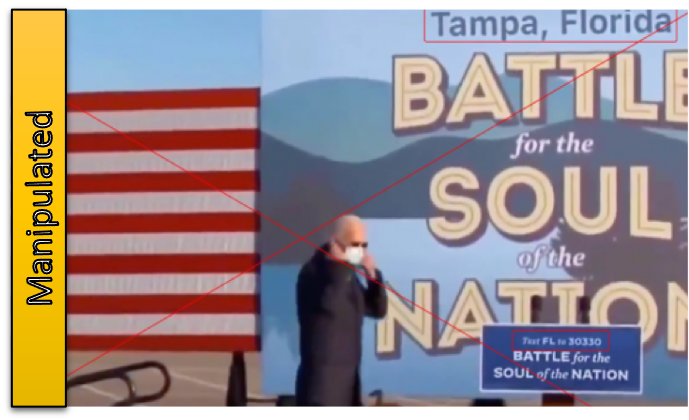}
\caption*{\textbf{(b2)} The clip above is altered by changing the location and the signs on the podium to a different US state, Florida.}
\end{subfigure}
\hfill
\rulesep
\begin{subfigure}{0.32\textwidth}
\includegraphics[width=\linewidth]{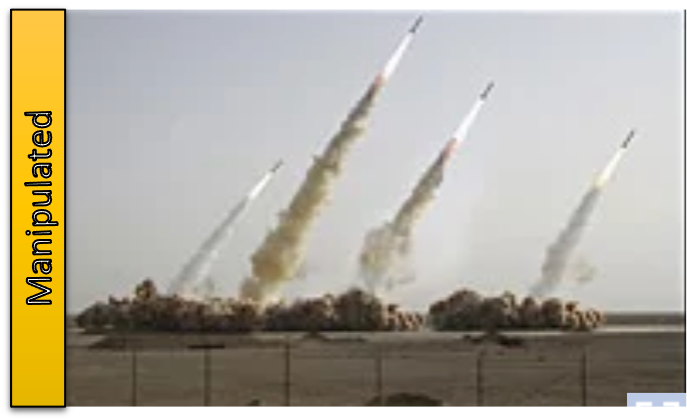}
\caption*{\textbf{(c2)} In an altered image released on Iran's Revolutionary Guards website, claimed that $4$ missiles were launched simultaneously.}

\end{subfigure}
\caption{\textbf{Spatial manipulations:}  \textbf{(a)}~\cite{FoxNewsr60online}, \textbf{(b)}~\cite{Falsevid34online}, and \textbf{(c)}~\cite{PhotoOfI88online} are examples of videos on social media spatially manipulated with the intent to mislead audiences.}
\label{fig:motivation}
\end{adjustbox}
\end{figure*}


%% file: tables/summary.tex
\begin{table*}[t]
    \centering
    \resizebox{\textwidth}{!}{
    \begin{threeparttable}
    \begin{tabular}{crcccccccccc}
     \toprule
     \textbf{Faces}&\textbf{Datasets}&\textbf{Release Date}&\multicolumn{2}{c}{\# \textbf{Videos}} &\multicolumn{2}{c}{\textbf{Source}} & \textbf{Attacks} & \textbf{Human} &  \textbf{Context} & \multicolumn{2}{c}{\textbf{Modality}} \\
     \cmidrule{4-7}
     \cmidrule{11-12}
     &&&Real&Fake&Original&Manipulated&&\textbf{Density}&&Visual&Audio\\
    \midrule
    \multirow{12}{*}{Only} &UADFV~\cite{UADFV}\Tstrut&Nov-18&$49$&$49$&YouTube&Deep Learning&$3$&\cellcolor{red!25}$1$&\cellcolor{red!25} \xmark&\cellcolor{green!25} \cmark&\cellcolor{red!25} \xmark\\
     
     &DF-TIMIT~\cite{DeepFake-TIMIT} &	Dec-18 &	$640$ & $320$&VidTIMIT~\cite{sanderson2002vidtimit} & Deep Learning& $3,4$&\cellcolor{red!25}$1$&\cellcolor{red!25} \xmark&\cellcolor{green!25} \cmark&\cellcolor{green!25} \cmark\\
     
     &FaceForensics++~\cite{faceforensics++}&Jan-19&$1000$&$4000$&YouTube&Deep Learning&$3,4$&\cellcolor{red!25}$1$&\cellcolor{red!25} \xmark&\cellcolor{green!25} \cmark&\cellcolor{red!25} \xmark\\
     
     &DFD~\cite{google}&	Sep-19&	$0$&	$3000$&	YouTube&	Deep Learning&$3$&\cellcolor{red!25}$1$&\cellcolor{red!25} \xmark&\cellcolor{green!25} \cmark&\cellcolor{red!25} \xmark\\
     
     &CelebDF~\cite{Celeb-DF}&	Nov-19&	$5907$&	$5639$&	YouTube&	Deep Learning&$3$&\cellcolor{red!25}$1$&\cellcolor{red!25} \xmark&\cellcolor{green!25} \cmark&\cellcolor{red!25} \xmark\\
     
     &DFDC~\cite{DFDC}&	Oct-21&	$23654$&	$104,500$&	Actors&	Unknown&$3$&\cellcolor{red!25}$1$&\cellcolor{red!25} \xmark&\cellcolor{green!25} \cmark&\cellcolor{red!25} \xmark\\
     
     &DeeperForensics 1.0~\cite{deeperforensics}&	Jan-21&	$50,000$&	$10,000$&	Actors&	Deep Learning&$3$&\cellcolor{red!25}$1$&\cellcolor{red!25} \xmark&\cellcolor{green!25} \cmark&\cellcolor{red!25} \xmark\\
     
     &WildDeepFake~\cite{zi2020wilddeepfake} &Jan-21&$3,805$	&$3,509$&Internet	&Internet	&$3,4,5$&\cellcolor{red!25}$1$&\cellcolor{red!25} \xmark&\cellcolor{green!25} \cmark&\cellcolor{red!25} \xmark\\
     
     &KoDF~\cite{kwon2021kodf}&	Aug-21&	$62,166$&	$175,776$&Actors	&	Deep Learning&$3,4,5$&\cellcolor{red!25}$1$&\cellcolor{red!25} \xmark&\cellcolor{green!25} \cmark&\cellcolor{green!25} \cmark\\
     
     &FakeAVCeleb~\cite{khalid2021fakeavceleb}&	Sep-21&	$490+$	&$20,000+$&VoxCeleb2~\cite{chung2018voxceleb2}	&Deep Learning	&$3,4$&\cellcolor{red!25}$1$&\cellcolor{red!25} \xmark&\cellcolor{green!25} \cmark&\cellcolor{green!25} \cmark\\
     
     &ForgeryNet~\cite{he2021forgerynet}&July-21	&	$91,630$	&$121,617$&	Multiple&Deep Learning	&$3,4$&\cellcolor{red!25}$1$&\cellcolor{red!25} \xmark&\cellcolor{green!25} \cmark&\cellcolor{green!25} \cmark\\
     
     &SR-DF~\cite{wang2021m2tr} &Apr-21	&	$1,000$	&$4,000$&YouTube	&	Deep Learning&$3,4$&\cellcolor{red!25}$1$&\cellcolor{red!25} \xmark&\cellcolor{green!25} \cmark&\cellcolor{green!25} \cmark\\
     
     &Khelifi et al.~\cite{khelifi2017perceptual} &	Jan-19&	$200$&	$200$&Multiple&	User Generated	&$6,7$&\cellcolor{red!25}$1$&\cellcolor{red!25} \xmark&\cellcolor{green!25} \cmark&\cellcolor{red!25} \xmark\\
     
     \cmidrule{1-12}
     \Tstrut
     \multirow{4}{*}{Beyond}&MTVFD~\cite{al2016development}&	2016&	$30$&	$30$&	YouTube&	User Generated&	$1,2$&	\cellcolor{red!25} $\leq1$	&\cellcolor{green!25} \cmark&	\cellcolor{red!25} \xmark&\cellcolor{red!25} \xmark\\
   
     &Liao et al~\cite{liao2013video} &	2013 & $10$ & $8$ &	Multiple & User Generated &	$1$ & \cellcolor{red!25} $\leq1$	& \cellcolor{green!25} \cmark & \cellcolor{green!25}\cmark & \cellcolor{red!25} \xmark\\
  
     &Su et al~\cite{su2015video} & 2015 & $7$ & $7$ & SONY DSCP10 & User Generated & $1$ &	\cellcolor{red!25} $\leq1$ &	\cellcolor{green!25} \cmark &\cellcolor{green!25} \cmark&	\cellcolor{red!25} \xmark\\
  
     \cmidrule{2-12}
     &\textbf{Ours}&	Nov-21&	$413$	&$413$&	Online Videos&User Generated	&\cellcolor{green!25}  &\cellcolor{green!25}upto $40$&\cellcolor{green!25} \cmark& \cellcolor{green!25} \cmark&\cellcolor{green!25} \cmark\\
     \bottomrule
    \end{tabular}
\end{threeparttable}
    }
    \caption{\textbf{Characteristics of Video Manipulation Datasets: }We compare \dataname~with state-of-the-art video manipulation datasets.}
    \label{tab:summary}
\end{table*}

%% file: tables/attacks.tex
\begin{table*}[b]
    \centering
    \resizebox{\textwidth}{!}{
    \begin{tabular}{lclll}
    \toprule
         & \textbf{S.No.} & \textbf{Attack} & \textbf{Method/Software} & \textbf{Description} \\
         \midrule
         \multirow{8}{*}{\rotatebox{0}{\textbf{Spatial}}} & \multirow{2}{*}{$1$} & \multirow{2}{*}{Copy-Move and Splicing}& \multirow{2}{*}{Adobe Photoshop$^\textbf{TM}$,  AfterEffects$^\textbf{TM}$} & Select and copy region within same video and paste \\
         
          & & &  & this somewhere else within the same video or different video\\
         \cmidrule{2-5}
         & $2$ & Retouching/Lighting & Adobe Lightroom $^\textbf{TM}$ & Brightness increase/decrease, Contrast Increase/Decrease, Median Filter\\
          \cmidrule{2-5}

         & \multirow{2}{*}{$3$} & \multirow{2}{*}{Face Swapping~(FS)} & FakeApp, FaceSwap~\cite{korshunova2017fast}&\multirow{2}{*}{Transferring a face from source to target image/video }\\
         
         && &  FaceShifter~\cite{li2019faceshifter}, FSGAN~\cite{nirkin2019fsgan}, DeepFaceLab~\cite{perov2020deepfacelab} &   \\
          \cmidrule{2-5}

         & \multirow{2}{*}{$4$} & \multirow{2}{*}{Face Re-enactment (FR)} & Neural Textures~\cite{thies2019deferred}, First-Order-Motion~\cite{siarohin2019first} & Using facial movements and expression deformations of\\
         
         &  &  &  Face2Face~\cite{thies2016face2face}, IcFace~\cite{tripathy2020icface}, FSGAN~\cite{nirkin2019fsgan}, & a face to guide the motions and deformations of another face\\
          \cmidrule{2-5}

         & $5$ & Audio-driven FR~(AFR) & Wav2Lip~\cite{prajwal2020lip}, APB2FACE~\cite{zhang2020apb2face}, ATFHP~\cite{yi2020audio} & Reenacting faces driven by a given audio signal to sync with lip movement\\
         
         \midrule
         
         \multirow{1}{*}{\rotatebox{0}{\textbf{Temporal}}} & $6$ & Temporal & Adobe Lightroom $^\textbf{TM}$ & Frame Dropping, Frame Insertion, Shifting in time, Frame Swapping\\
         
         \midrule
         
         \multirow{1}{*}{\rotatebox{0}{\textbf{Geometric}}} & $7$ & Geometric & Adobe Lightroom $^\textbf{TM}$ & Cropping, Resizing, Rotation, Shifting\\
         
        \bottomrule
    \end{tabular}
    }
    \caption{\textbf{Attacks: }We summarize the various attacks that have explored in prior literature for manipulating images and videos.}
    \label{tab:attacks}
\end{table*}

%% file: sections/2-relatedwork.tex
\begin{figure}[t]
    \centering
    \includegraphics[width =\columnwidth]{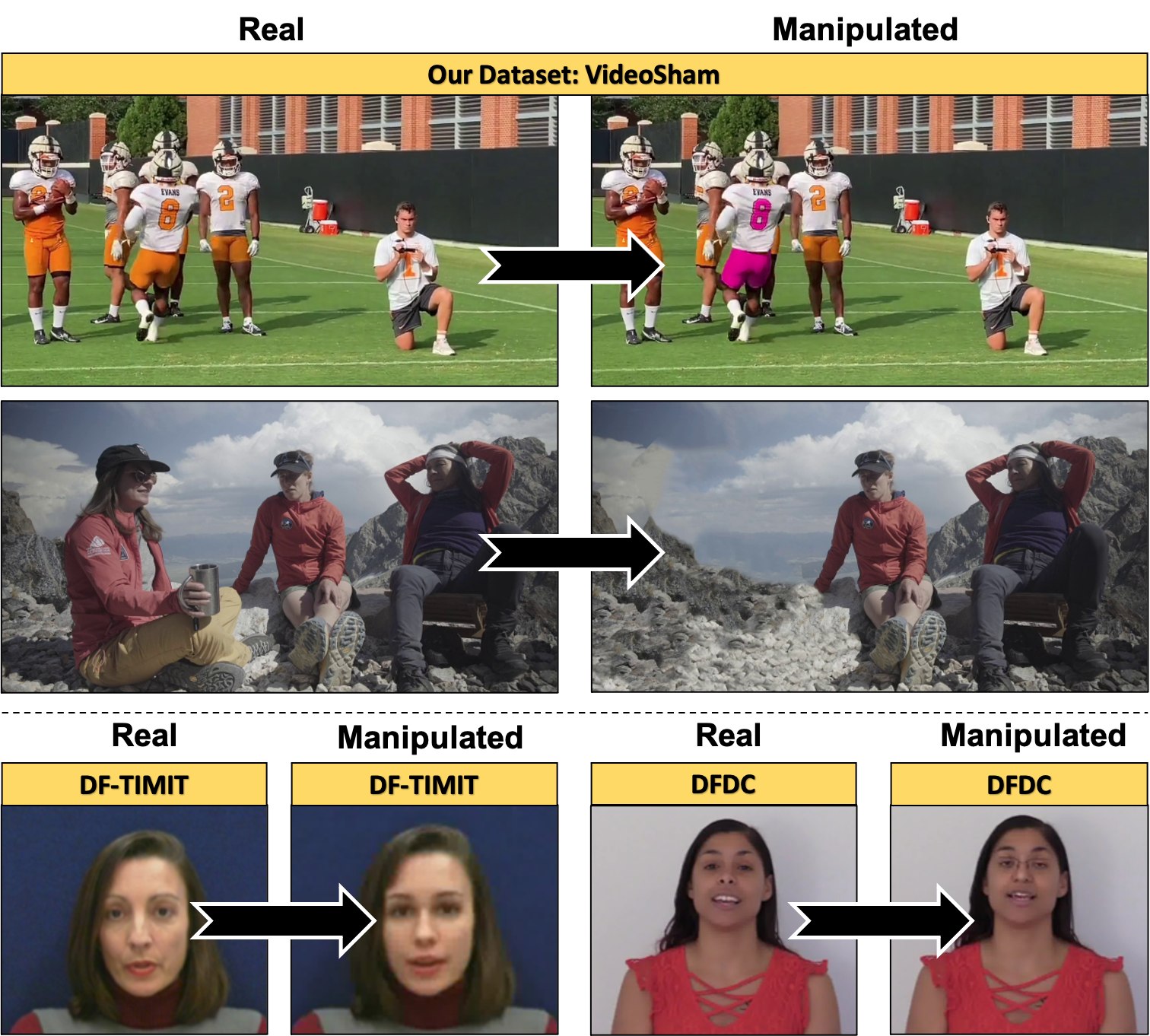}
     \caption{\small{\textbf{\dataname:} \textit{(top)} \dataname~consists of diverse, context-rich, and human-centric manipulated videos by professional video editors via $6$ spatial and temporal attacks (\textit{e.g.} jersey color change and person removal). \textit{(bottom)} In contrast, deepfake datasets (DF-TIMIT and DFDC) only consist of facial manipulations individual subjects from a close-up angle.}}
    \label{fig:cover}
\end{figure}
\section{Related Work}
\label{sec:relatedwork}
In this section, we discuss previous works in detection of manipulated and deceptive media content. To begin with, we first discuss the video manipulation techniques used to create such fake videos in Section~\ref{subsec:manipulation-attacks}. Then in Section~\ref{subsec:manipulation-datasets}, we summarize various datasets and benchmarks for video manipulations. We also survey different techniques used for detecting deepfake videos in Section~\ref{subsec:deepfake-detection-methods} and generic video forensic methods in Section~\ref{subsec:video-forensic-methods}. 
\subsection{Video Manipulation Techniques/Attacks}
\label{subsec:manipulation-attacks}
Manipulation techniques, or attacks, are broadly categorized as spatial~\cite{amerini2011sift}, temporal~\cite{khelifi2017perceptual}, and geometric~\cite{khelifi2017perceptual} in the literature (see Table~\ref{tab:attacks}). Basic examples of spatial attacks include copy-move and image/video splicing which correspond to spatially or temporally shifting an object to a different location in the same video or a different video, respectively. Retouching, another common attack, involves aesthetic edits like adjusting brightness, contrast, and other parameters of digital content. More recently, people have used AI to alter facial features to create deepfake videos. AI-based techniques are comprised of two major attack approaches, Face Swapping~\cite{nirkin2019fsgan,perov2020deepfacelab} and Face Re-enactment~\cite{thies2016face2face,thies2019deferred,siarohin2019first}. Face Swapping switches the subject's face with the face of another person and Face Re-enactment alters the subject's facial expressions. Temporal attacks involve swapping, duplicating, inserting, and deleting frames of video, giving the impression that the video has been sped up or slowed down. Finally, geometric attacks include operations like cropping and rotations.
\subsection{Video Manipulation Datasets}
\label{subsec:manipulation-datasets}
Creating benchmarks of video manipulations is a challenging task as this may require per-frame manipulations. Some of the datasets~(like Khelifi et al.~\cite{khelifi2017perceptual}, MTVFD~\cite{al2016development}, Liao et al.~\cite{liao2013video}, Su et al.~\cite{su2015video}, Media Forensics Challenge~\cite{guan2019mfc}) are very small in volume containing $7 - 200$ videos each, these datasets are also \textbf{not publicly available}. Most of these videos have $0$ or $1$ subjects present in the frame with very little background context. More recently, AI-synthesized attacks like \textit{face swapping}, \textit{face re-enactment}, and \textit{audio-driven face re-enactment} have led to the creation of datasets like UADFV~\cite{UADFV}, FaceForensics++~\cite{faceforensics++}, DeeperForensics1.0~\cite{deeperforensics}, WildDeepFake~\cite{zi2020wilddeepfake}. Because these datasets are generated using learning methods; some of these datasets have upto $100$k videos. However all of these datasets have strictly $1$ subject per video with the face being predominant part of the frame with no background context at all. Many datasets are missing audio except DFDC~\cite{DFDC}, DF-TIMIT~\cite{DeepFake-TIMIT}, KoDF~\cite{kwon2021kodf}, FakeAVCeleb~\cite{khalid2021fakeavceleb}, ForgeryNet~\cite{he2021forgerynet} and SR-DF~\cite{wang2021m2tr}. 
\subsection{Deepfake Detection Methods}
\label{subsec:deepfake-detection-methods}
The goal of the deepfake detection approaches is to algorithmically distinguish fake videos from real videos. A large portion of these methods are focused on detecting visual artifacts especially on the finer regions of the face~(like eyes, mouth and teeth~\cite{Matern2019ExploitingVA}). Some approaches specifically focus on abnormalities like inconsistent head pose orientations~\cite{yang2019exposing}, asynchronous lip movement and speech~\cite{haliassos2021lips} and unnatural eye blinking~\cite{li2019exposing}. Prior work have also observed and exploited the fact that temporal coherence is not enforced effectively in the synthesis process of deepfakes~\cite{temporal1,temporal2} and exploit this in detection methods. More recently, interesting affective computing approaches that focus on correlated emotion signals from audio-visual cues~\cite{mittal2020emotions,agarwal2020detecting}, and detecting signals like heart rate and breathing rate~\cite{qi2020deeprhythm} from the videos have also been proposed. However, it is clear that due to the nature of the datasets~(single-person, face-centered videos), these approaches focus only on facial cues and audio cues. 
\subsection{Video Forensic Methods}
\label{subsec:video-forensic-methods}
Developments in video forensics literature focus on two specific attacks; Copy-Move and Splicing~(Row 1 in Table~\ref{tab:attacks}) and Temporal attacks~(Row 6 in Table~\ref{tab:attacks}). Most conventional copy-move forgery detection methods mainly consist of three components~\cite{cozzolino2015efficient}: (1) feature extraction, (2) matching, and (3) post-processing. A variety of features have been explored, e.g., DCT (Discrete Cosine Transform)~\cite{mahmood2016copy}, DWT (Discrete Wavelet Transform) and KPCA (Kernel Principal Component Analysis)~\cite{bashar2010exploring}, Zernike moments~\cite{ryu2013rotation}. Consequently, some end-to-end deep learning based copy-move forgery detection methods were proposed~\cite{wu2018busternet,wu2018image,li2019exposing}. However these efforts are limited to images. Another interesting development, still in naive stages is deep learning methods to detect inpainting in videos~\cite{zhou2021deep}. Some of the methods in detecting temporal attacks~(also called as intra-frame manipulations) use the consistency of velocity field~\cite{wu2014exposing} and optical flow~\cite{wang2014video}. These methods can recognize frame insertion and frame deletion attacks. Similarly, Zhao et al.~\cite{zhao2018inter} use inter-frame similarity analysis to detect frame duplications in the videos. Finally, Long et al.~\cite{long2019coarse} propose a coarse-to-fine framework based on deep Convolutional Neural Networks~(CNN) to detect potential frame duplications. 

%% file: sections/3-dataset.tex
\section{Our Dataset- \dataname}
\label{sec:dataset}
In this section, we present details on the dataset creation process~(Section~\ref{subsec:creation}) and discuss some of the salient features and characteristics of \dataname~(Section~\ref{subsec:analysis}).
\subsection{Creation and Annotation Process}
\label{subsec:creation}
\subsubsection{Source Videos}
We have a total of $836$ videos comprising of $413$ original videos and $413$ manipulated versions, each corresponding to one of the original videos. We obtain our source videos from an online video website~(vimeo~\footnote{\href{www.vimeo.com}{www.vimeo.com.}}) and only include videos attributed with a CC-BY~(Creative Commons) license. In addition, we avoid videos with brands, children, objectionable content, TV show/movie clips and videos with copyrighted music. We trim these original videos to a specific length~(upto $5\mathrm{-}30$ seconds) before we perform any manipulation attack. 
\subsubsection{Manipulation Attacks}
We employ a total of $6$ manipulation attacks for creating our dataset. As per prior literature, we also categorize these attacks into spatial and temporal attacks\footnote{\footnotesize We do not use geometric attacks, as they have been shown to be easily detected.}. We visually show the distribution of the attacks in Figure~\ref{fig:attack-split}~(attacks outlined in blue are spatial attacks, outlined in pink are temporal attacks). We describe each of the attack below. 
\setdefaultleftmargin{.5cm}{}{}{}{}{}
\begin{enumerate}[(1)]
\item [$-$] ATTACK 1 (Adding an entity/subject): In this attack we select an entity or a subject from some other sources and place them in the current video. This attack is somewhat similar to copy-move attack. 
\item [$-$] ATTACK 2 (Removing an entity/subject): In this attack, we basically select an entity or a subject in the video and remove it from all the frames and fill in the gap with background settings. To do this, we used  content-aware fill in Adobe AfterEffects$^{\text{TM}}$ and some deep learning methods for generating masks~\cite{he2017mask} and performing video inpainting~\cite{kim2019deep,kim2020vipami}.
\item [$-$]  ATTACK 3 (Background/Color Change): We focus on a particular aspect of the video, and change the background of the video, or color of a small entity in the video. 
\item [$-$] ATTACK 4 (Text Replaced/Added):  We perform edits like adding some text in the video or removing or replacing already existing text in the video. 
\item [$-$] ATTACK 5 (Frames Duplication / Removal/ Dropping): This attack is specifically to render the video temporally inconsistent. We choose to perform one of these manipulations, randomly duplicating frames, removing or dropping frames in the video. This also includes slowing down a video. 
\item [$-$] ATTACK 6 (Audio Replaced): Audio modality is a very important aspect for videos. To manipulate this, we replace the existing audio with some other audio. 
\end{enumerate}
We visually depict the $4$ spatial attacks~(ATTACK 1, ATTACK 2, ATTACK 3, and ATTACK 4) in Figure~\ref{fig:qual}. 
\subsubsection{Manipulated Videos}
We worked with $3$ professional video editors hired on Upwork~\footnote{\href{www.upwork.com}{www.upwork.com.}}. The editors were shortlisted based on their experience and were well-versed with Adobe AfterEffects$^{\text{TM}}$, the software used for creating these edits. Each editor was assigned tasks, i.e. source videos, start and end timestamp to be edited and a one-line description of the manipulation to be performed. We provide all videos and the attacks performed for every video.
\input{images/graphs}

\noindent\textbf{Dataset Analysis:}
In Figure~\ref{fig:attack-split}, we present the distribution of attacks for the $413$ videos, each lasting $1-31$ seconds. The average length of videos in our dataset is around $8$ seconds long.  We also run an object detection model~~\footnote{\href{https://github.com/roboflow-ai}{https://github.com/roboflow-ai.}} to count the number of people/agents in every video~(Figure~\ref{fig:no-of-people}). More than $80\%$ of the videos in our dataset contains at least one subject.

%% file: images/graphs.tex
\begin{figure*}[t]
\centering
  \begin{subfigure}[h]{.32\textwidth}
    \includegraphics[width=\linewidth]{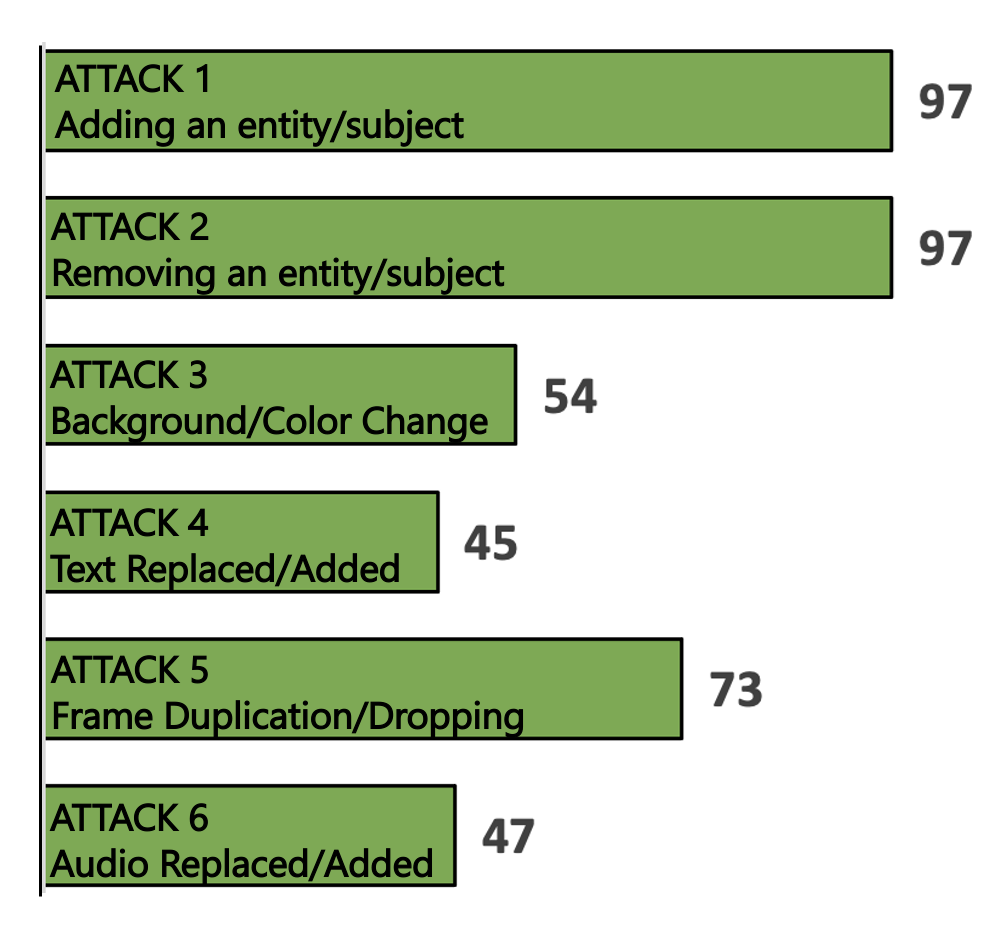}
    \caption{\textbf{Attack distribution:} Distribution of videos that are attacked with different manipulation techniques. Attacks $1-4$ are spatial attacks, and Attacks $5-6$ are temporal attacks.}
    \label{fig:attack-split}
  \end{subfigure}
  \quad
  \begin{subfigure}[h]{.29\textwidth}
    \includegraphics[width=\linewidth]{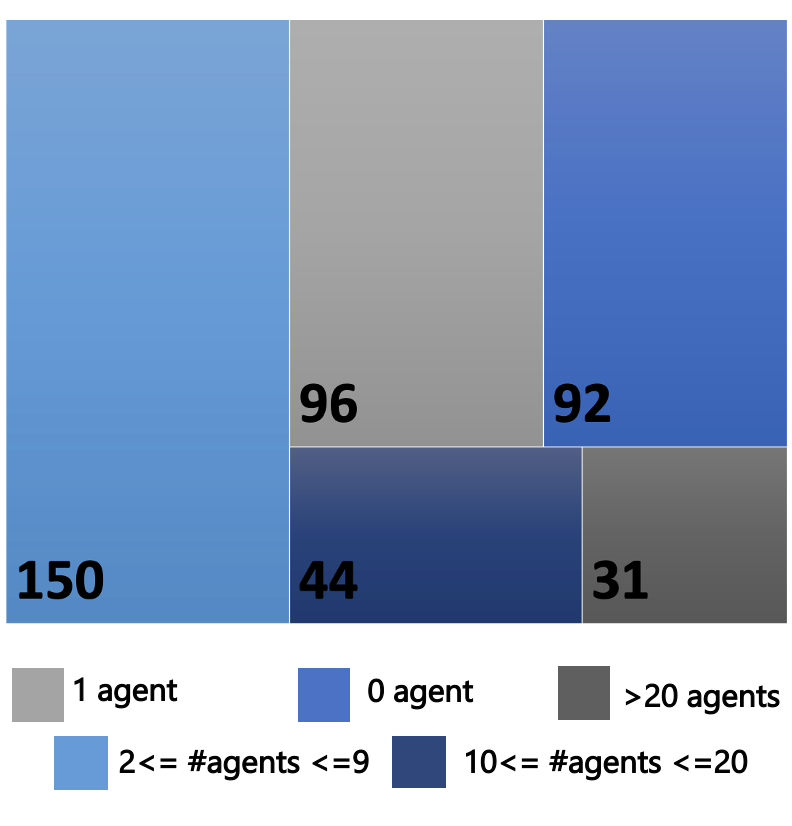}
    \caption{\textbf{Density distribution:} Distribution of videos according to number of persons present in each video. This is considerably high w.r.t. the existing datasets.}
    \label{fig:no-of-people}
 \end{subfigure}
 \quad
 \begin{subfigure}[h]{.3\textwidth}
    \includegraphics[width=\linewidth]{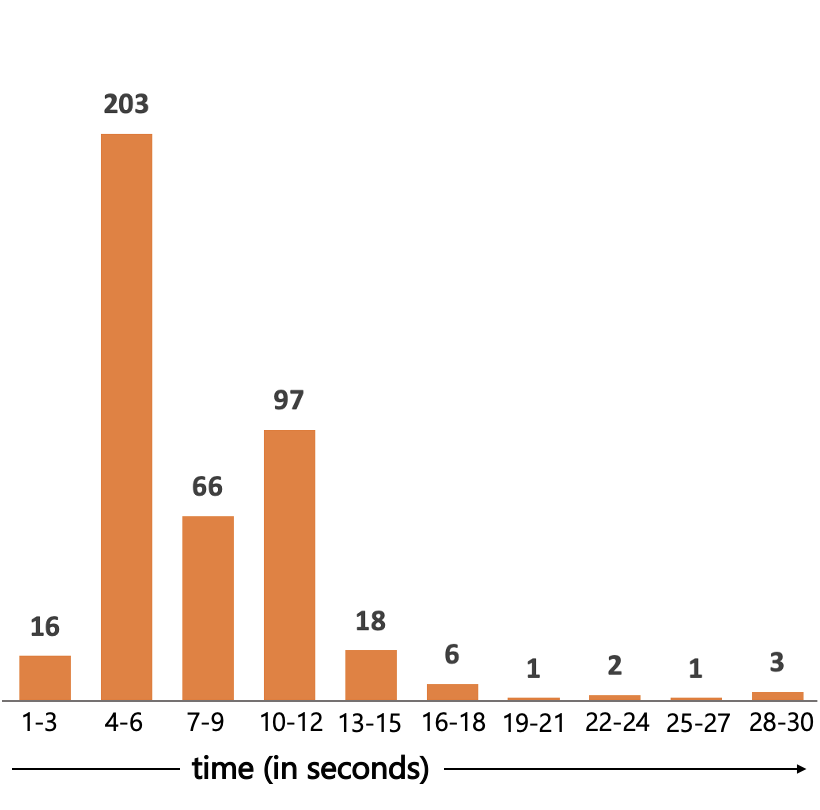}
    \caption{\textbf{Duration distribution:} Distribution of videos according to duration or length of each video (in seconds). The average length of our videos is $8$ seconds.}
    \label{fig:length}
  \end{subfigure}
\caption{\textbf{Dataset statistics:} We visually present various statistics for \dataname for better insights.}
 \label{fig:graphs}
\end{figure*}

%% file: sections/4-experiments.tex
\section{Experiments and Results}
\label{sec:experiments}
We elaborate on three experiments we perform to highlight the importance, novelty and usecase of \dataname. To begin with, we present the analysis of how well humans fair in detecting these attacks in Section~\ref{subsec:expt1}, followed by analysis of the performance of state-of-the-art deepfake detection methods and video forensic techniques in Section~\ref{subsec:expt2}. Finally in Section~\ref{subsec:expt3}, we present some ideas and preliminary results for using interdisciplinary ideas for detecting such attacks. 
\subsection{Expt 1: How Well Do Humans Perform?}
\label{subsec:expt1}

\begin{figure}[t]
    \centering
    \includegraphics[width=\columnwidth]{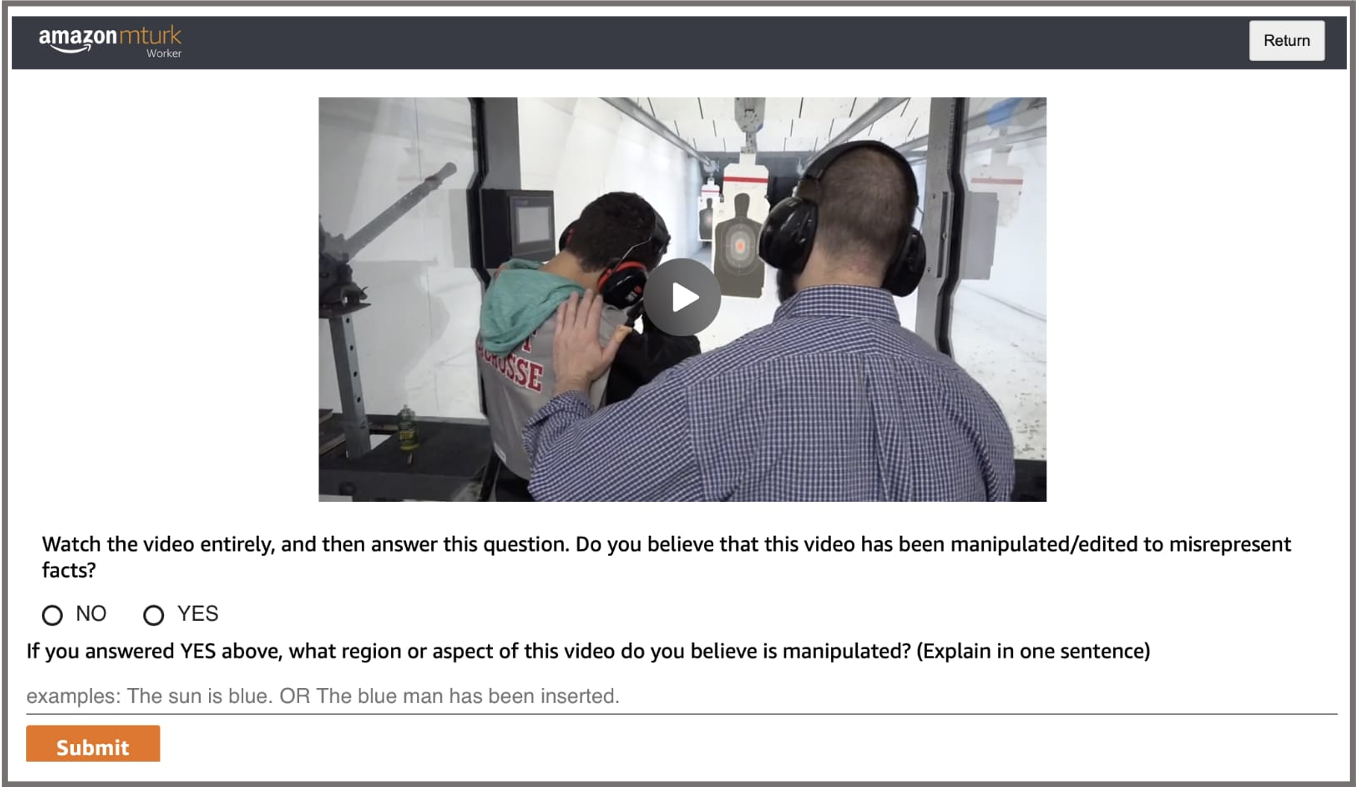}
    \caption{\textbf{User Study Setup: }We present the Amazon Mechanical Turk setup used (Section~\ref{subsec:expt1}). }
    \label{fig:user_study_fig}
\end{figure}

\textbf{Setup: }We first shortlist $60$ videos from \dataname. Out of these, $30$ videos are real and the remaining $30$ are manipulated~($5$ videos per attack). We recruit human participants from Amazon Mechanical Turk (AMT) and show each video to $20$ participants. The participants are requested to watch the full video; followed by two questions. In the first question, the participants are asked to respond to the following prompt in either a yes or no - ``Do you believe this video has been manipulated/edited to misrepresent facts?''. And, in the second question we ask them to explain in a sentence what they felt was manipulated with the following prompt- ``If you answered YES above, what region or aspect of this video, do you believe is manipulated.''. Note that participants are not informed whether videos are manipulated or not. They are also not informed about the set of attacks. We show the setup in Figure~\ref{fig:user_study_fig} which was used to collect a total of $1200$ responses from AMT participants.

\textbf{Study Analysis: }We summarize the responses of the user study in Table~\ref{tab:user_study_table}. Both the real and manipulated videos receive $600$ responses each. We observe that out of the $600$ responses~(corresponding to the $30$ real videos), $342$, i.e., $57\%$ were correctly identified as real. Similarly, out of $600$ responses for the manipulated videos, $389$ were incorrectly identified as real, i.e., $35.2\%$ of these responses correctly identified manipulated videos. 
Analyzing the responses by the type of attack, we observe that human participants are able to identify $45\%$ of the videos manipulated using ATTACK $6$. For the other attack types, the proportion of manipulated videos labeled as `fake' ranges from $13$-$31$\%. Furthermore, we notice that human participants are able to more successfully identify manipulated videos that are modified using temporal attacks~(ATTACK $5$ and ATTACK $6$) than spatial attacks~(ATTACK $1 -$ ATTACK $4$). Moreover, we also received some number of responses from participants explaining their rationale behind reporting a manipulated video. From the responses received, there is no clear evidence that suggests that participants are able to identify the manipulated region/kind in case of spatial edits. But, they were somewhat able to correctly identify the manipulated edit in case of temporal attacks. This would imply that a subset of our selection of attacks are indiscernible to the human eye. 

\underline{Statistical Tests:} Next we consider statistical tests to see if humans are able to tell a real video from a manipulated video. We consider the following quantities for this test, define $p_1 = P(\text{declaring video real} | \text{real video})$. Also, let $p_2= P(\text{declaring video real} | \text{manipulated video})$. If humans are able to tell real videos apart from manipulated videos, we expect $p_1$ to be larger that $p_2$. Hence, we test the one-sided statistical hypothesis:
\begin{equation*}
    H_0: p_1 = p_2 \quad \text{against} \quad H_1: p_1 > p_2.
\end{equation*}
We test this hypothesis with the test statistic $(p_1-p_2)$\footnote{Given our sample size, we have a $86\%$ statistical power of detecting a difference if the true values are $p_1 = 0.75$ and $p_2 =0.74$.}. In Table~\ref{tab:user_study_table} we present the difference of proportions as well as the one-sided $p-$value of the test for each attack type. The first thing to note is that when combining across all attack types (last row), we see that even though $p_1$ is slightly bigger than $p_2$, this difference is not statistically significant ($p-$value of $0.177$). This suggests that our edits are not discernible to human evaluators. When we break it down by attack type, we observe that only for ATTACK $6$ (audio replacement), humans are more likely to declare such edits as manipulations ($p-$value $<0.001$). For ATTACK $4$ (text replaced or added), there is weak statistical evidence of humans detecting this manipulation ($p-$value of $0.097$). For all other attacks, there is no statistical evidence that humans can tell when a video has been manipulated using that strategy. It is particularly telling that when an entity/subject is added or removed (ATTACKs $1$ and $2$), more of our human subjects declare such manipulated videos as real than they declare unedited videos. This shows how modern editing tools can be used to manipulate videos in a way that humans have no way of telling such edits just by looking at the video. This observation establishes the need to build high quality video manipulation detection algorithms that can label manipulated videos at scale.  
\input{tables/userstudy}

\subsection{Expt 2: How Well Do Machines Perform?}
\label{subsec:expt2}
To answer this question better, we evaluate state-of-the-art deepfake detection methods and video forensics techniques on \dataname. 
\input{tables/DF_forensics}

\noindent\textbf{Deepfake Detection Methods: }We evaluate Li et al.~\cite{li2019exposing}, XceptionNet~\cite{faceforensics++} and Mittal et al.~\cite{mittal2020emotions} on \dataname. Deepfake videos generated using data-driven methods can only synthesize face images of a fixed size, and they must undergo an affine warping to match the configuration of the source’s face. Due to resolution inconsistencies between warped face and background context, there are various artifacts on the synthesized faces. Li et al.~\cite{li2019exposing} detects such artifacts by comparing the generated face areas and their surrounding regions with a dedicated Convolutional Neural Network (CNN) model. On the other hand, XceptionNet~\cite{faceforensics++} is a transfer learning model which is also a CNN architecture, which was originally trained for the classical object detection task and later finetuned for deepfake detection on FaceForensics++ dataset. Finally, Mittal et al. propose an approach that simultaneously exploits the audio (speech) and video (face) modalities and also the perceived emotion features extracted from both the modalities to detect any falsification or alteration in the input video. They use the correlation between the modalities to detect a fake video.
\input{tables/prelim-results}

\noindent\textbf{Video Forensics Techniques: }We evaluate Long et al.~\cite{long2019coarse} and Liu et al.~\cite{liu2021two} on \dataname. Both of these methods are state-of-the-art methods in video forensics literature. While, Long et al.~\cite{long2019coarse} is specifically for detecting cases of frame duplications in a video, Liu et al.~\cite{liu2021two} specifically focus on detecting copy-move attacks. For all the methods, we use pretrained models and report the results when evaluated on \dataname in Table~\ref{tab:evaluation}. 
\begin{figure}[t]
    \centering
    \includegraphics[width=\linewidth]{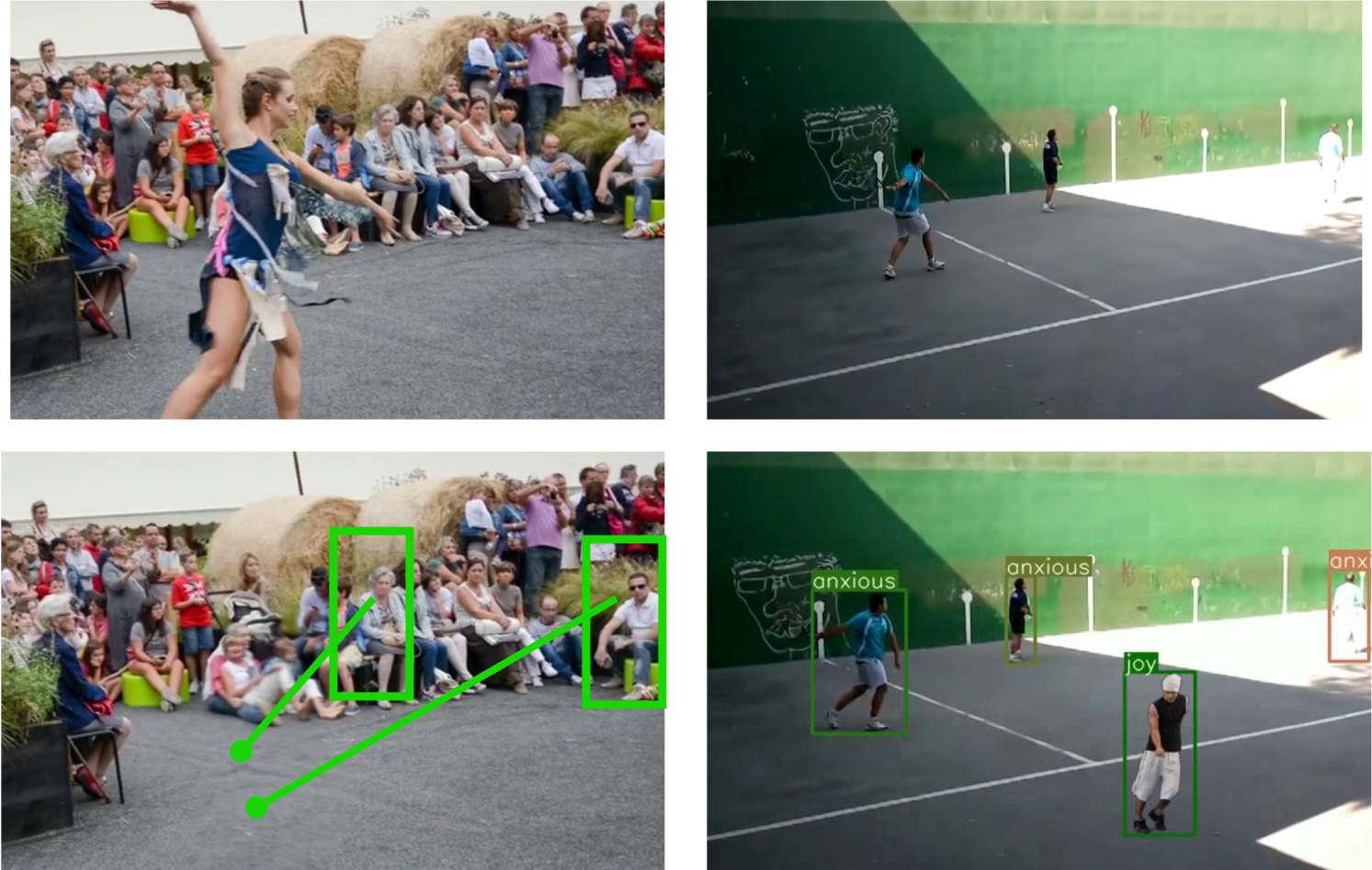}
    \caption{\textbf{Inter-Agent Dynamics and Multimodal Ideas for Detecting Manipulations: }We show the output of the automated techniques used to identify manipulated videos in \dataname. \textit{(Column $1$)} In the first column, we remove the main subject from the foreground. We identify this as a manipulated image using a gaze tracking algorithm by noting that there is no object at the location of the crowds gaze direction.  \textit{(Column $2$)}Here, we manipulate an image by inserting the man in black shirt. We use emotion recognition techniques to infer that this false subject has an affective state that is not in tune to those of the other players.}
    \label{fig:ideas-image}
\end{figure}

\noindent\textbf{Study Analysis: }All the $5$ shortlisted methods are less then $50\%$ accurate on \dataname. This is understandable, as all the deepfake methods~(Li et al.~\cite{li2019exposing}, MesoNet~\cite{Mesonet}, and Mittal et al.~\cite{mittal2020emotions}) are trained specifically to look for manipulations in faces. Moreover, these method are not used to inferencing on videos with more or less than $1$ person in the frame and with so much context information. Hence, we observe that these methods are only inferring based on artifacts caught near the face regions in the \dataname~videos. We also observe that, Mittal et al. specifically are able to detect some of the temporal manipulations well; which is because the method is trained to look for correlation between audio and visual modalities. Similarly, even the video forensics techniques are specifically performing well on attacks that they have been trained for, i.e. ATTACK $5$ for Long et al. and ATTACK $1$ and ATTACK $2$ for Liu et al.~\cite{Liu2016DeepSH}. ATTACK $3$~(color change) and ATTACK $4$~(text replacement) tend to remain hard to be detected by most of these methods. 

\subsection{Expt 3: Beyond DeepFake Detection and Video Forensic Techniques}
\label{subsec:expt3}
One can observe from the experiments in the previous section, that all the methods are largely dependent on the visual artifacts. However, given the diversity of attacks used to manipulate videos, we hypothesize the use of inter-agent and multimodal analysis models for detecting such manipulations. We show preliminary results in Figure~\ref{fig:ideas-image}.

\noindent \textbf{\ul{Strategy 1 (Gaze):}} To begin with, we believe that tracking gaze of subjects can be useful for detection experiments. Gaze following is a task in computer vision to identify objects and regions that the subject of interest is focusing on. The idea behind this strategy is to identify manipulated images by using gaze following to locate ``absent'' targets and/or ``out-of-context'' subjects in the video.  
To perform some preliminary analysis we deploy GazeFollow~\cite{recasens2016they}. More specifically, for each frame, we begin by obtain the spatial coordinates of the subject's head's bounding box and pass this information as input to the gaze tracking algorithm, GazeFollow~\cite{recasens2016they}, which outputs the location of the subject's gaze. The final step in this strategy is to run an object detector to obtain a confidence score $c_g$ corresponding to an object present at the gaze location. A low confidence score indicates a manipulated frame.

\noindent \textbf{\ul{Strategy 2 (Affect):}} In this strategy we propose the use of affective cues. When we track and look for affective disparities in affective state of different subjects. Prior works in psychology~\cite{Kleinsmith2013AffectiveBE} and empirical works~\cite{Mittal2020EmotiConCM} that subjects in social settings often share affective states. 
We use facial expressions, body postures and scene understanding to perceive the affective states of all subjects. We use the model EmotiCon~\cite{Mittal2020EmotiConCM} trained on EMOTIC dataset~\cite{kosti2017emotic} to perceive these affective states and obtain an affective confidence score $c_{a}$. By empirically assigning a threshold, $\tau$ on the two confidence scores, we flag a video as manipulated. We observe that these two techniques help detect ATTACK 1 and ATTACK 2 significantly well. We add quantitative results for the same in Table~\ref{tab:prelim-quant-results}. We show two qualitative results of these ideas in Figure~\ref{fig:ideas-image}. 

Experiment $3$ shows that, in addition to human assessment, specialized deepfake detection techniques, and video forensics, other approaches that are not intended for identifying manipulated videos can be used. 

%% file: tables/userstudy.tex

\begin{table}[t]
    \centering
\resizebox{\columnwidth}{!}{
  \begin{tabular}{p{1pt}p{1pt}p{1pt}ccccccccc}
\toprule[1.5pt]
&&\multicolumn{10}{c}{(A) $1200 \ \textrm{\textit{responses}} \ (20 \ \textrm{participants} \times (30 \ \textrm{real} + 30 \ \textrm{manipulated videos}))$ }\hfill \\
\toprule
\multirow{2}{*}{\textbf{GT}}&&&\textbf{\#Resp}&\textbf{Rep} &\textbf{Rep} &\multirow{2}{*}{$\bm{p_1}$}&\multirow{2}{*}{$\bm{p_2}$}& \multirow{2}{*}{$\bm{p_1 - p_2}$}&\multirow{2}{*}{$\bm{p-}$ \textbf{value}}&\multirow{2}{*}{\textbf{CI(l)}}&\multirow{2}{*}{\textbf{CI(u)}}\\
&&&\textbf{(Total)}&\textbf{Real} & \textbf{Fake}&& &  &&& \\
\midrule
\multirow{2}{*}{\rotatebox{90}{\textbf{Real}}}\Tstrut &&& \multirow{2}{*}{$600$} &\multirow{2}{*}{$454$} & \multirow{2}{*}{$146$}&\multirow{2}{*}{$0.757$}&& \multirow{2}{*}{$0$}&\multirow{2}{*}{$-$}&\multirow{2}{*}{$-$}&\multirow{2}{*}{$-$}\\
&&&&&&&&&&&\\
\cmidrule{3-12}
\multirow{6}{*}{\rotatebox{90}{\textbf{Manipulated}}}&\multirow{6}{*}{\rotatebox{90}{\textbf{Attack}}}& \Tstrut 1& $100$ &$87 $&$13 $&$0.757$ &$0.87$&$-0.113$&$0.991$&$-0.182$&$1$\\
& &2& $100$ & $79 $ & $21 $&$0.757$&$0.79$&$-0.033$&$0.725$&$-0.112$&$1$\\
&& 3& $100$  &$74 $ & $26 $&$0.757$&$0.74$&$0.016$&$0.408$&$-0.066$&$1$\\
&& 4& $100$  &$69 $ & $31 $&$0.757$&$0.69$&$0.066$&$0.097$&$-0.020$&$1$\\
&& 5& $100$ & $75 $ & $25 $&$0.757$&$0.75$&$0.006$&$0.493$&$-0.076$&$1$\\
&& 6& $100$ & $55 $ & $45 $&$0.757$&$0.55$&$0.207$&$\bm{<0.001}$&$0.114$&$1$\\
\cmidrule{4-12}
&\Tstrut &&$600$& $439$ & $161$ &$0.757$&$0.732$&$0.0250$&$0.177$&$-0.018$&$1$\\
\bottomrule[1.5pt]
\end{tabular}
}
\caption{\textbf{Human Performance: } We observe that participants are unable to detect ATTACK 3~($26\%$) and ATTACK 4~($31\%$). Videos manipulated using ATTACK 5 are relatively easier to detect ($75\%$). }%
\label{tab:user_study_table}
\end{table}

%% file: tables/DF_forensics.tex
\begin{table}[t]
\centering
\resizebox{\linewidth}{!}{
\begin{tabular}{cccc|cc|cc|cc|cc|cc}
\toprule[1.25pt]
&&&&\multicolumn{10}{c}{\textbf{Predicted}} \Tstrut\\
\cmidrule[1.05pt]{5-14}
\multirow{3}{*}{\textbf{GT}}&&&\multirow{3}{*}{\#Vid.}&\multicolumn{6}{c|}{Deepfake Detection Methods}\Tstrut&\multicolumn{4}{c}{Video Forensics Techniques}\\
\cmidrule[1.05pt]{5-14}
&&&&\multicolumn{2}{c}{Li et al~\cite{li2019exposing}}\Tstrut&\multicolumn{2}{c}{MesoNet~\cite{Mesonet}}&\multicolumn{2}{c|}{Mittal et al.~\cite{mittal2020emotions}}&\multicolumn{2}{c}{Long et al.~\cite{long2019coarse}}&\multicolumn{2}{c}{Liu et al.~\cite{liu2021two}}\\
\cmidrule[1.05pt]{5-14}

&&&&\multicolumn{2}{c}{\textbf{Predicted}}&\multicolumn{2}{c}{\textbf{Predicted}}&\multicolumn{2}{c|}{\textbf{Predicted}}&\multicolumn{2}{c}{\textbf{Predicted}}&\multicolumn{2}{c}{\textbf{Predicted}}\Tstrut\\
&&&&Real\Tstrut&Fake&Real&Fake&Real&Fake&Real&Fake&Real&Fake\\
\midrule
\multirow{2}{*}{\rotatebox{90}{\textbf{Real}}}\Tstrut && & \multirow{2}{*}{$413$} & \multirow{2}{*}{$188$} & \multirow{2}{*}{$225$} & \multirow{2}{*}{$167$} & \multirow{2}{*}{$246$} & \multirow{2}{*}{$238$} & \multirow{2}{*}{$175$} & \multirow{2}{*}{$219$} & \multirow{2}{*}{$194$} & \multirow{2}{*}{$234$} & \multirow{2}{*}{$179$}\\
& &&  &  &  &  &&  && &  &  &  \\

\midrule
\multirow{6}{*}{\rotatebox{90}{\textbf{Manipulated}}}
& \multirow{6}{*}{\rotatebox{90}{\textbf{Attack}}}&1\Tstrut & $97$ & $76$ & $21$ & $93$ & $4$ & $92$ & $5$ & $86$ & $11$ & $68$ & $29$ \\
& &2 & $97$ & $63$ & $34$ & $84$ & $13$ & $66$ & $31$ & $84$ & $13$ & $46$ & $51$ \\
& &3 & $54$ & $35$ & $19$ & $37$ & $17$ & $49$ & $5$ & $50$ & $4$ & $38$ & $16$ \\
& &4 & $45$ & $32$ & $13$ & $34$ & $11$ & $42$ & $3$ & $39$ & $6$ & $34$ & $11$ \\
& &5 & $73$ & $70$ & $3$ & $67$ & $6$ & $54$ & $19$ & $25$ & $48$ & $68$ & $5$ \\
& &6 & $47$ & $45$ & $2$ & $44$ & $3$ & $31$ & $16$ & $38$ & $9$ & $41$ & $6$ \\
\cmidrule{2-14}
&& & $413$ \Tstrut & $321$ & $92$ & $359$ & $54$ & $334$ & $79$ & $322$ & $91$ & $295$ & $118$\\
\bottomrule[1.25pt]
\end{tabular}
}
\caption{\textbf{Machine Performance: } We evaluate $3$ state-of-the-art deepfake detection methods and $2$ video forensics techniques on \dataname. It is apparent that these algorithms do not perform well on \dataname, speaking to its complexity and diversity.}
\label{tab:evaluation}
\end{table}

%% file: tables/prelim-results.tex
\begin{table}[H]
\centering
\resizebox{\linewidth}{!}
{
\begin{tabular}{cccc}
\toprule[1.5pt]
\textbf{GroundTruth} & \textbf{\# videos}&\textbf{Reported} & \textbf{Reported} \\
&&\textbf{Real} & \textbf{Manipulated} \\
\midrule
Real & $413$ & $286$ & $127$\\
Attack 1 & $97$ & $32$ & $65$\\
Attack 2 & $97$ & $35$ & $62$\\
\bottomrule[1.5pt]
\end{tabular}
}
\caption{\textbf{Quantitative Results~(Expt 3):} For some preliminary analysis, we explore two ideas, \textit{gaze} and \textit{affect} of all agents involved. We observe that these two ideas in itself can effectively detect manipulations of the kind, ATTACK $1$ and ATTACK $2$.}
\label{tab:prelim-quant-results}
\end{table}

%% file: sections/7-discussion.tex
\section{Conclusion and Future Directions }
\label{sec:discussions}
Our goal with the expt 1~(Section~\ref{subsec:expt1}) and expt 2~(Section~\ref{subsec:expt2}) was to understand how well humans can detect some of the manipulations that occur today circulated on social media. We also wanted to understand if the developments in the deepfake detection and video forensic literature match up to these manipulation attacks. Finally, through expt 3~(Section~\ref{subsec:expt3}) we want to propagate the idea of using ideas beyond detection of visual artifacts for scalable models for video manipulation detection. 

We conclude from expt 1~(Section~\ref{subsec:expt1}) and expt 2~(Section~\ref{subsec:expt2}) that both humans and machines~(5 methods shortlisted) struggle to detect these manipulations successfully. We believe that these are attacks of concern, as they are going undetected even by human participants. Moreover, we emphasize that these manipulations play a big role in many real-world video manipulations~(Figure~\mbox{\ref{fig:motivation}}). 

More generally, we believe that computer vision algorithms perform almost comparable to humans in most of these ATTACKS. However, most methods are very attack-specific and do not generalize well to other attacks. Mostly every deepfake detection method fails to handle videos with more than $1$ subject and hence have a very limited scope. Also, importantly most of the deepfake detection methods require huge amounts of training samples; and this is not a realistic assumption. It is important to build methods which can be less computationally intensive and at the same time are also able to generalize well. Similarly, methods in video forensics also are only able to handle very specific attacks. These are less dependent on data, but computationally expensive as they are more or less, inference based methods. 

We believe following are some knowledge gaps and research agendas that can help the society combat the increasing problem of misinformation, frauds and cybercrimes occurring due to manipulated media content shared online.
\begin{enumerate}
    \item There is a need to build detection models focused on more diverse attacks or video manipulations. Through \dataname, we attempted to include some of the attacks that have not been studied before owing to a lack of a dataset. We hope this dataset can be a step towards achieving better detection models for all the $6$ attacks. 
    \item Moreover it is important to increase the scope of detection ideas being used currently for detecting manipulations. Current methods are extremely focused on visual perception. Our goal through experiment 3 was to show through very preliminary analysis that ideas based on inter-agent dynamics and multimodal cues can be a promising literature source. Another promising idea, is to include domain knowledge in detecting manipulations; as humans we have some contextual information which the detection models severely suffer from.
    \item Largely all existing methods require a significant amount of training data to train the models. But, with newer manipulations and attacks on videos, it will become impossible to keep up with detection models for the same. We need to reduce the dependence on training data build detection models that are as generalizable as possible to potential attacks. 
\end{enumerate}
\section{Ethical Considerations}
We note that our dataset sources videos from an online video website that are attributed with a CC-BY license, and we do not retain any metadata corresponding to the creators of the videos. In addition, we do not collect any personal information of the human participants in the subsequent user study conducted on AMT. We expect that our dataset is an effort towards mitigating and fighting against malicious manipulations of online digital content.

\section*{Acknowledgements}

This research is partially supported by ARO Grant W911NF2110026.

%% file: sections/appendix.tex
\section{\dataname: Qualitative Examples}
We present some qualitative examples from our dataset here. We show one example from each of the $4$ spatial attacks in this picture. 
\begin{figure*}[b!]
    \centering
    \includegraphics[width=\textwidth]{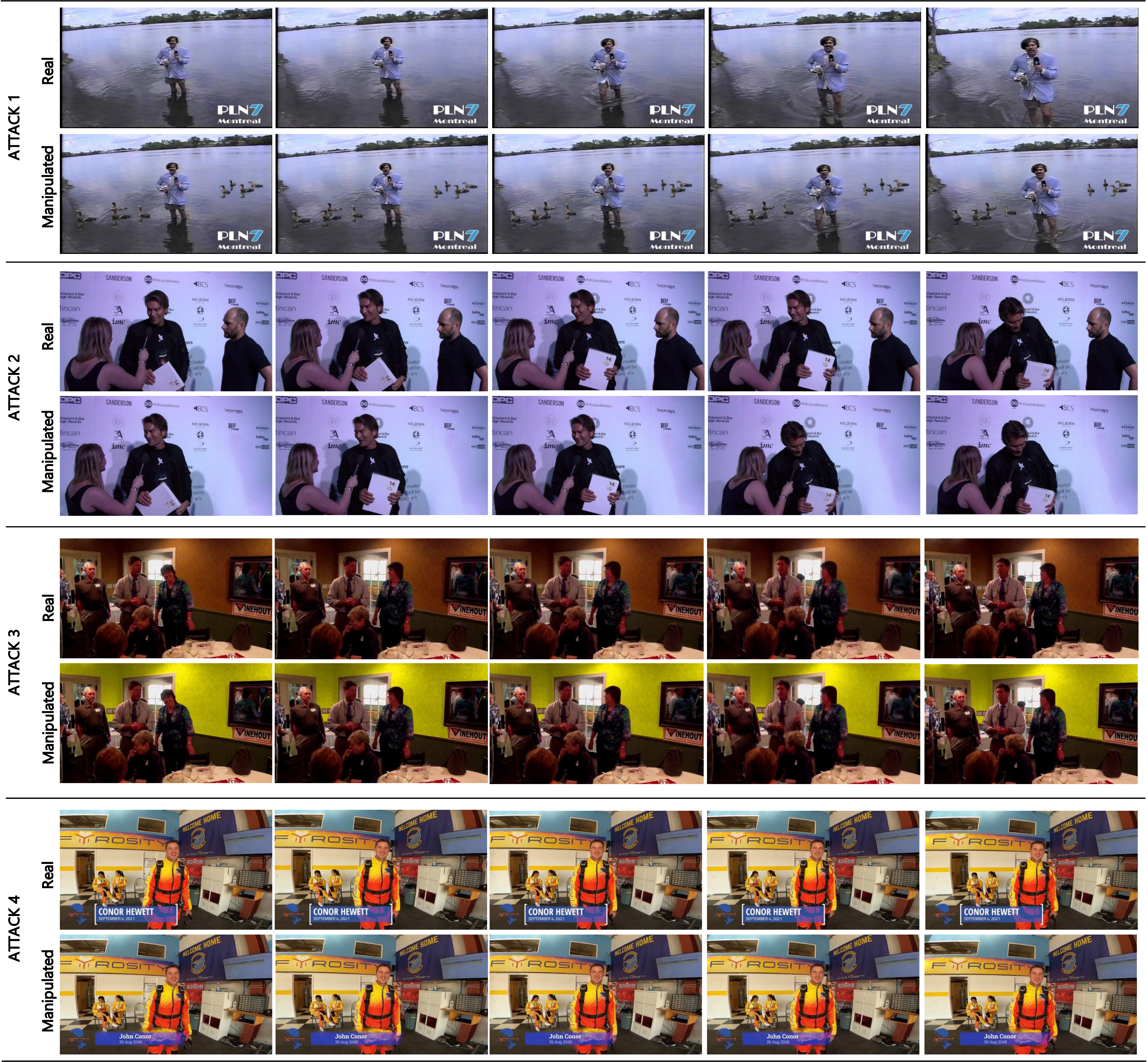}
    \caption{\small{\textbf{\dataname~Examples: }We present a series of frames, both for real and manipulated videos for the $4$ \textit{spatial} attacks. In ATTACK 1, we add ducks in the water behind the person talking in the microphone. In ATTACK 2, we remove the man in the black shirt to the right corner. In ATTACK 3, we change the color of the walls to yellow. And, finally in ATTACK 4, we alter the name of the person talking. We were not able to add examples of the $2$ \textit{temporal} attacks~(ATTACK 5, ATTACK 6) here.}}
    \label{fig:qual}
\end{figure*}
\